\documentclass[runningheads]{llncs}
\usepackage[T1]{fontenc}
\usepackage{graphicx}
\usepackage{subcaption}

\usepackage{algorithm}
\usepackage{algpseudocode}

\begin{document}
\title{Obtaining Example-Based Explanations\\from Deep Neural Networks}
\author{Genghua Dong\inst{1}\orcidID{0009-0004-4494-2320} \and
Henrik Boström\inst{1}\orcidID{0000-0001-8382-0300} \and
Michalis Vazirgiannis\inst{1,2}\orcidID{0000-0001-5923-4440} \and
Roman Bresson\inst{1}\orcidID{0009-0004-7500-4900}}
\authorrunning{Dong et al.}
\institute{KTH Royal Institute of Technology, Stockholm, Sweden
\email{\{genghua,bostromh,mvaz,bresson\}@kth.se}\and
LIX, École Polytechnique, IP Paris, France\\
\email{mvazirg@lix.polytechnique.fr}}
\maketitle              
\begin{abstract}
Most techniques for explainable machine learning focus on feature attribution, i.e., values are assigned to the features such that their sum equals the prediction.
Example attribution is another form of explanation that assigns weights to the training examples, such that their scalar product with the labels equals the prediction. The latter may provide valuable complementary information to feature attribution, in particular in cases where the features are not easily interpretable. Current example-based explanation techniques have targeted a few model types only, such as k-nearest neighbors and random forests. In this work, a technique for obtaining example-based explanations from deep neural networks (EBE-DNN) is proposed. The basic idea is to use the deep neural network to obtain an embedding, which is employed by a k-nearest neighbor classifier to form a prediction; the example attribution can hence straightforwardly be derived from the latter. Results from an empirical investigation show that EBE-DNN can provide highly concentrated example attributions, i.e., the predictions can be explained with few training examples, without reducing accuracy compared to the original deep neural network. Another important finding from the empirical investigation is that the choice of layer to use for the embeddings may have a large impact on the resulting accuracy.

\end{abstract}
\section{Introduction}
\begin{figure}[h]
\includegraphics[width=\textwidth]{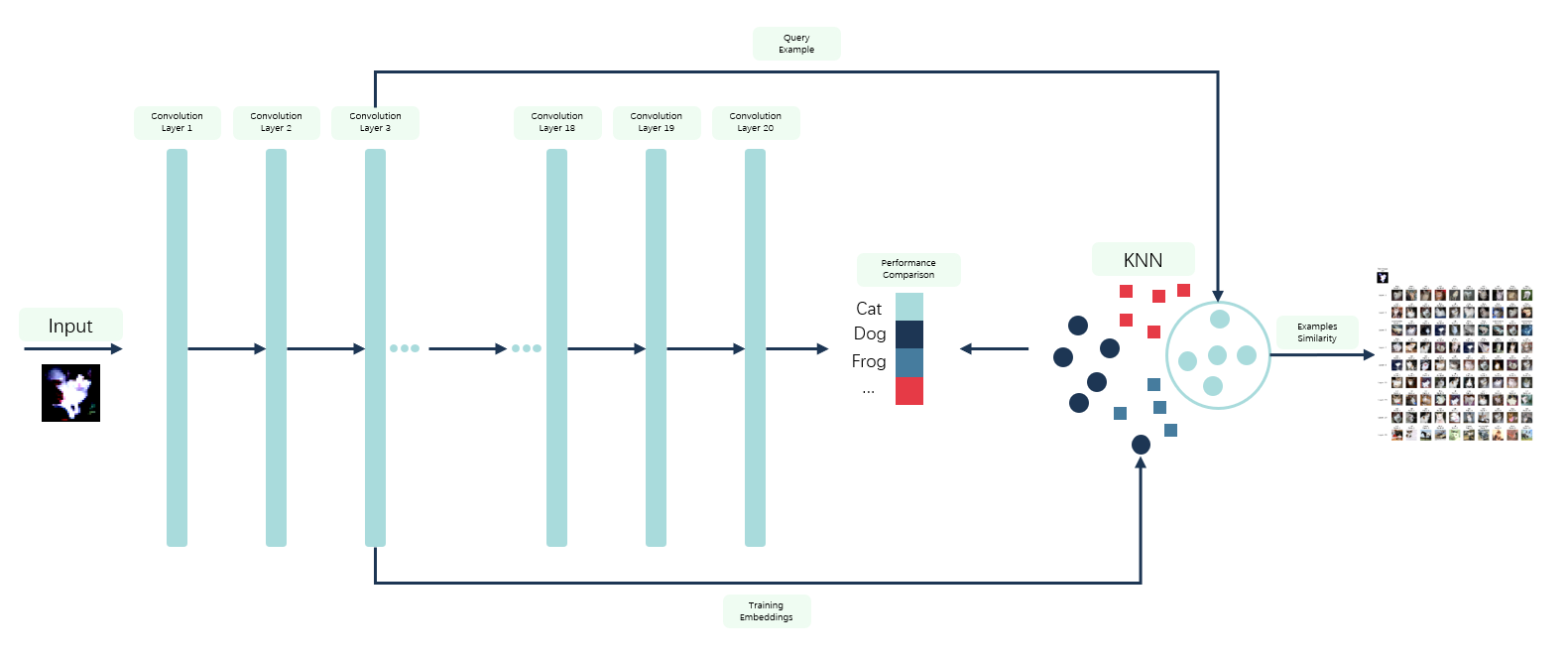}
\caption{The structure of EBE-DNN} \label{ebeknnstructure}
\end{figure}
Deep neural networks (DNN) play an indispensable role in modern society, including in areas such as autonomous driving, AI-assisted medical diagnosis, and financial fraud prevention. However, due to their 'black box' nature, there has been growing concern about the inner workings of these networks \cite{gunning2019darpa}. Explainable artificial intelligence (XAI)\footnote{The term XAI is in this work used as a synonym to explainable machine learning.} aims to produce models for which the predictions can be explained in a way that is understandable to humans and that may provide insights into the inner workings of the models, with the end goal of allowing users to trust and manage them effectively. XAI provides transparency into how decisions are made by a machine learning model, identifying the key features or examples contributing to its predictions \cite{gunning2019darpa,adadi2018peeking}.

Some machine learning models are by design interpretable, e.g., logistic regression, decision trees, decision rules, and k-nearest neighbors (KNN). Others, such as DNN, can be interpreted by employing post-hoc analysis techniques, using e.g., feature-based, concept-based, rule-based, and example-based explanation methods \cite{molnar2020interpretable}. Feature-based explanations (FBE) describe how input features contribute to the model output. This class includes the well-known model-agnostic techniques LIME \cite{ribeiro2016should} and SHAP \cite{lundberg2017unified}, as well as the model-specific Grad-CAM \cite{selvaraju2020grad}, which targets Convolutional Neural Network (CNN)-based models for computer vision tasks. Since enumerating all feature values involved in a prediction is often not useful, concept-based approaches, such as TCAV \cite{kim2018interpretability}, address these limitations by finding the concept, e.g., a color and an object, embedded within the latent space learned by DNN. Anchors \cite{ribeiro2018anchors} is a representative from the class of rule-based explanation method, which explains individual predictions of any black box classification model by finding a decision rule that 'anchors' the prediction sufficiently.

In contrast to the aforementioned techniques, example-based explanations (EBE) enable an understanding of the model behavior through specific examples in the training data \cite{molnar2020interpretable}. Central to this is the concept of example attribution, which refers to identifying and quantifying the influence of training examples on a model’s prediction. Example attribution provides a direct connection between the model's predictions and its training data, offering insights into how specific examples guide the model's decision-making. A straightforward method for example attribution is the KNN algorithm, which bases predictions on the most common class among the nearest neighbors of an example. However, its reliance on raw data without feature extraction limits its performance in tasks requiring high-level feature understanding, such as CIFAR-10, where features like edges or textures are critical. Beyond KNN, methods for example attribution in other models, such as random forests, assign weights to training examples based on their contributions to predictions \cite{bostrom2024example}. However, research exploring example attribution in the context of DNN remains sparse.

Our focus is on forming the example attribution of a test example, which involves identifying and presenting training examples that influence the model's prediction.  Recent surveys provide different categorizations of EBE methods, see e.g., \cite{poche2023natural}. However, these methods often face limitations in practical deep learning tasks. For instance, the prototype and criticisms approach \cite{kim2016examples} focuses on representing the data distribution by selecting prototypes that are representative of all the data points and criticisms that are not well represented by the prototypes. However, this method does not directly provide example-level attributions for individual predictions, limiting its utility for explaining specific test examples. Similarly, counterfactual explanations \cite{guidotti2024counterfactual} consist of slight modifications to the test instance such that the prediction is affected. While it may provide some insights, this technique relies on the generation of semantically meaningful counterfactuals, which is challenging and often requires prior knowledge.

To address these limitations, we propose EBE-DNN, a method that leverages the strengths of both KNN and DNN to form example attributions for test examples. As illustrated in Fig.~\ref{ebeknnstructure}, embeddings from a trained DNN are used as the input for KNN, enabling the retrieval of training examples that are most common to the test image in terms of the DNN’s learned feature representation. By combining the interpretability of KNN-based example attributions with the feature extraction ability of DNN, EBE-DNN bridges the gap between explanation and model performance. It maintains the accuracy of the underlying DNN while providing example-based explanations. This approach ensures that example attributions are not only aligned with the model's decision-making process but also provide support for the predictions in the form of examples, enabling trust and transparency in deep learning models. Our method differs from previous work such as Deep k-Nearest Neighbors (DkNN) \cite{papernot2018deep}, which employs KNN in multiple DNN layers to estimate prediction credibility and detect out-of-distribution samples. In contrast, EBE-DNN is explicitly designed to improve interpretability by retrieving specific training examples that contribute to a test example’s prediction, aiming for highly concentrated example attributions rather than global estimates of prediction confidence.

The remainder of this paper is structured as follows. In section \ref{EBEKNN}, we introduce details of EBE-DNN. In Section \ref{EmpiricalInvestigation}, we illustrate and evaluate EBE-DNN on image classification tasks. Finally, we summarize the main conclusions and discuss future work in section \ref{conclusion}.

\section{EBE-DNN}\label{EBEKNN}

The proposed approach, EBE-DNN, which is outlined in algorithm \ref{alg:EBE-DNN}, operates by leveraging embeddings extracted from a specific layer of a deep neural network (DNN) to generate example attributions and predict the label of a test example. Given a set of training examples $X$ with their corresponding labels $y$, and a deep neural network $D$, embeddings are generated with respect to a specified layer $l$, upper bounded by $D_{L}$ (the number of layers in $D$), resulting in a transformed set of examples $X'$. The transformation step aims to convert the high-dimensional input examples to a more useful feature space, reflecting both low-level and high-level properties. Similarly, the test example $X_{n+1}$ is converted into its embedding $X_{n+1}'$ using the same transformation function. The $k$ nearest neighbors of the test instance in the transformed space are retrieved, using some distance metric ($\delta$). From these examples, which in the algorithm are denoted as ${X'{(1)}, \ldots, X'{(k)}}$, where each $(i)$ corresponds to the index of the example with rank $i$, the resulting example attribution for the test instance are obtained, i.e., $N = {(X_{(1)}, y_{(1)}), \ldots, (X_{(k)}, y_{(k)})}$. Finally, the predicted label for the test instance is given by the mode of the labels from the retrieved examples. The algorithm implicitly assumes uniform weighting of the $k$ returned examples, i.e., each weight is $1/k$, but alternative weighting schemes are possible.


{\centering
\begin{minipage}{.7\linewidth}
\begin{algorithm}[H]
\caption{EBE-DNN}\label{alg:EBE-DNN}
\begin{algorithmic}[1]
\Require
\Statex $X = \{X_1, \ldots, X_n\}$ \Comment{objects}
\Statex $y = \{y_1, \ldots, y_n\}$ \Comment{labels}
\Statex $D$ \Comment{deep neural network}
\Statex $0 < k \leq n$ \Comment{no. of examples}
\Statex $0 < l \leq D_{L}$ \Comment{layer no.}
\Statex $\delta$ \Comment{distance metric (e.g.,cosine)}
\Statex $X_{n+1}$ \Comment{test object}
\State $X' \gets Transform(X, D, l)$
\State $X_{n+1}' \gets Transform(X_{n+1}, D, l)$
\State $\{X'_{(1)}, \ldots, X'_{(k)}\} \gets KNN(X_{n+1}', X', k, \delta)$
\State $N \gets \{(X_{(1)}, y_{(1)}), \ldots, (X_{(k)}, y_{(k)})\}$
\State $\hat{y} \gets Mode(\{y_{(1)}, \ldots, y_{(k)}\})$
\\
\Return $\hat{y}, N$
\end{algorithmic}
\end{algorithm}
\end{minipage}
\par
}
\vspace{0.5cm}

\section{Empirical Investigation}\label{EmpiricalInvestigation}
In this section, we investigate the example attributions generated by EBE-DNN and evaluate its predictive performance across three widely studied image classification tasks. Our primary objective is to demonstrate that EBE-DNN effectively provides example attributions for test images without compromising the predictive accuracy of the employed DNN.

\subsection{Experimental Setup}

We evaluate EBE-DNN using three widely adopted datasets: MNIST \cite{lecun1998mnist}, Fashion-MNIST \cite{xiao2017fashion}, and CIFAR-10 \cite{krizhevsky2009learning}. MNIST is a dataset of grayscale $28 \times 28$ images of handwritten digits (0–9), while Fashion-MNIST consists of grayscale $28 \times 28$ images of clothing items, such as T-shirts, pants, and shoes. CIFAR-10 is a more diverse dataset containing grayscale $36 \times 36$ images of 10 classes, including animals (e.g., cats, dogs, horses) and vehicles (e.g., cars, trucks, airplanes). MNIST and Fashion-MNIST each have 60~000 training images and 10~000 test images, while CIFAR-10 contains 50~000 training images and 10~000 test images.

We select ResNet18 \cite{he2016deep} as the backbone model for EBE-DNN due to its proven effectiveness in image processing tasks. ResNet18 consists of 4 blocks and 20 convolutional layers, which are sequentially numbered from shallow to deep layers for reference. Among these, layers 8, 13, and 18 perform downsampling, enabling the network to reduce spatial dimensions. After training ResNet18 on each dataset using PyTorch \cite{paszke2019pytorch} with a batch size of 128, a learning rate of 0.001, and 60 epochs, the model weights are frozen to ensure consistency during embedding extraction.

To generate example attributions, embeddings are obtained from each convolutional layer of ResNet18 for both training and test images. These embeddings, representing the semantic features learned by ResNet18, differ across layers due to the hierarchical nature of feature abstraction in networks. To investigate the impact of the choice of layer, we apply EBE-DNN to each layer, i.e., $l \in \{1, \ldots, 20\}$. The function $KNN$ in the algorithm is using the Scikit-learn implementation \cite{pedregosa2011scikit}, with cosine similarity as the distance metric. This metric was chosen as it is particularly well-suited for high-dimensional spaces and is robust to magnitude scaling introduced by operations like convolution, pooling, and batch normalization. 


For illustration, we first present some example attributions provided by EBE-DNN. To investigate whether the example attribution comes at a cost in terms of predictive performance (accuracy), EBE-DNN is compared to the underlying DNN (ResNet18).

\subsection{Illustration of Example Attributions}\label{ExamplesAttribution}

In this section, we present example attributions generated by EBE-DNN for two CIFAR-10 test images, a cat and an automobile, demonstrating how the method identifies training examples that contribute to the model's predictions, here using $k = 10$ neighbors and embeddings from layer 14 of ResNet18. In addition, we explore how example attributions vary across different layers for the automobile test image, showing how the supporting training examples may change with the choice of convolutional layer.


\begin{figure}[tbp]
    \centering
    \begin{subfigure}[b]{1.0\textwidth}
        \centering
        \includegraphics[width=\textwidth]{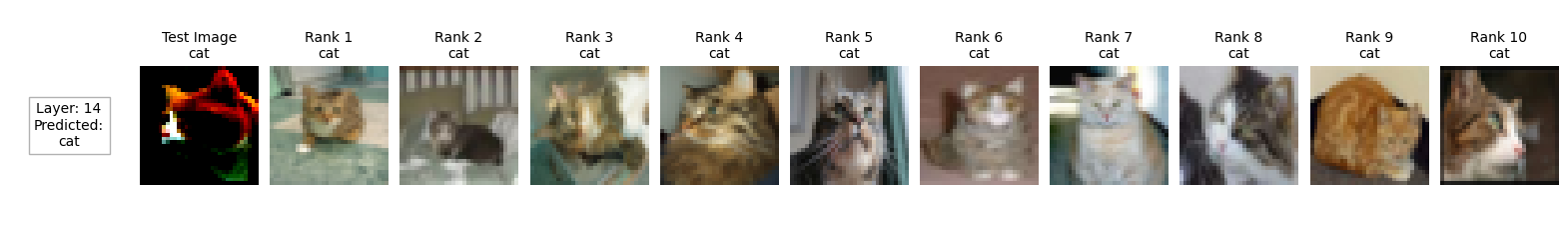} 
        \caption{Example attribution of a cat image using convolutional layer 14}
        \label{cat14}
    \end{subfigure}
    
    \vspace{1em}    
    
    \begin{subfigure}[b]{1.0\textwidth}
        \centering
        \includegraphics[width=\textwidth]{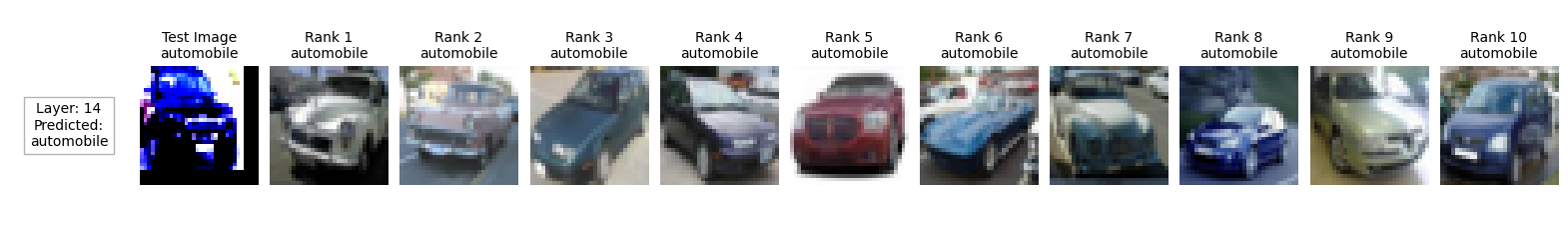}
        \caption{Example attribution of an automobile image using convolutional layer 14}
        \label{automobile14}
    \end{subfigure}
    \caption{Example attributions of cat and automobile test images}
    \label{layer14}
\end{figure}

\subsubsection{Example Attribution for a Cat Image} In Fig.~\ref{cat14}, the example attribution for a test image of a cat is shown, with all the training examples that contribute to a prediction being ranked. By inspecting the retrieved examples, some key properties such as fur patterns, facial structure, and ear shape, can be identified, e.g., most of the retrieved images show cats with a white patch of fur around their mouths or chests. Additionally, the round facial shapes in the examples suggest that the model relies on structural similarity in facial features to classify the image. The triangular ear shapes observed across all examples further underscore their importance as a distinguishing property. While ranks 4 and 9 deviate slightly by lacking prominent white fur, other properties seem to be quite consistent across all ten examples.

\subsubsection{Example Attribution for an Automobile Image} In Fig.~\ref{automobile14}, the example attribution for a test image of an automobile is shown. The retrieved examples indicate that the test image is classified as an automobile based on properties such as the shape of the car’s front face, the visibility of tires, and color similarity. A majority of the retrieved examples capture automobiles with a front face structure similar to that of the test image, with headlights positioned symmetrically and contours of the car hood reflecting the design of the test automobile. Although the examples ranked 2, 3, 10 deviate slightly with smaller car hoods influenced by shooting angles, they still maintain structural consistency with the test image. The presence of one or two visible tires across all examples further indicate their role as a critical property for classifying automobiles. Additionally, six of the retrieved images share a blue color similar to the test image, suggesting that color consistency contributes to classification but plays a secondary role compared to structural properties. This example attribution gives an indication of the ability of ResNet18 to capture visual and structural properties essential for classifying automobiles.

\begin{figure}
\includegraphics[width=\textwidth]{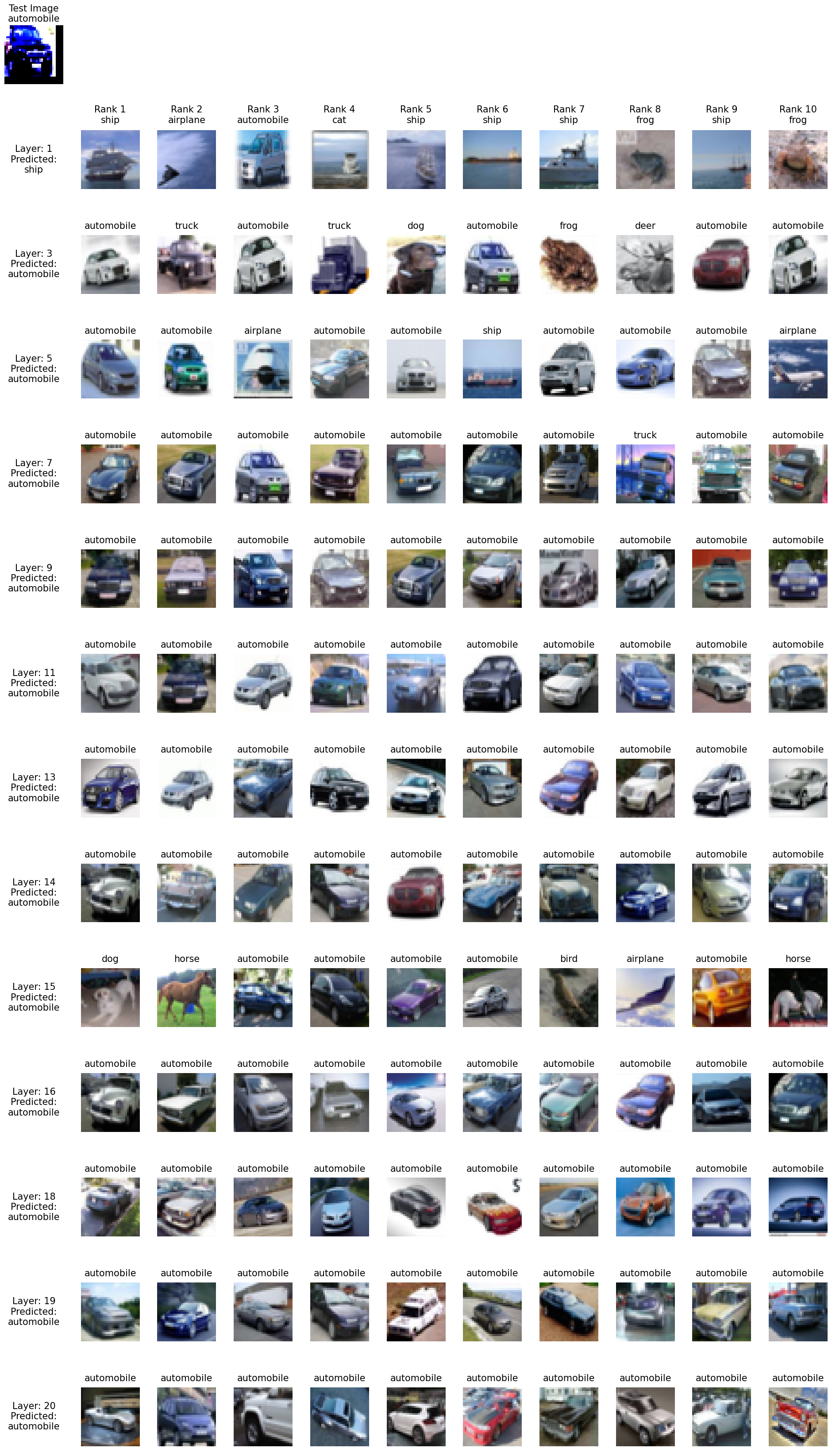}
\caption{Example attributions of an automobile test image from CIFAR-10} \label{figCIFAR-10automobilesim}
\end{figure}

\subsubsection{Example Attribution Across Layers}
In Fig.~\ref{figCIFAR-10automobilesim}, example attributions for the above automobile test image are presented, using embeddings from different convolutional layers of ResNet18.

In the top three layers, the retrieved images primarily reflect low-level properties, such as color and texture. For instance, many of the retrieved examples share the blue-and-white color scheme of the test image, which corresponds to the car’s blue body and white background. Additionally, texture patterns, such as the smooth surfaces of the car body, are captured prominently. However, the retrieved images exhibit minimal structural consistency, as the embeddings at these layers focus on localized patterns rather than overall object structures. Notably, higher-ranked images (e.g., rank 1 and 2) generally show better alignment with the test image’s color distribution and texture patterns compared to lower-ranked images, which display greater variability in these properties.

As the embeddings progress through middle layers, such as layers 7, 9, and 14, the retrieved examples begin to incorporate more high-level and discriminative properties. Properties like the car hood’s shape, the placement of headlights, and the presence of visible tires become more prominent. These layers achieve a balance by preserving low-level details, such as color and texture, while simultaneously capturing structural features critical for classification. The rank order in this stage reflects the relevance of structural consistency: higher-ranked images exhibit more similar frontal shapes and component placements to the test image, while lower-ranked images still align structurally but may vary slightly in less significant features, such as the angle of the car hood or the positioning of tires.

In the deeper layers, such as layers 18, 19, and 20, the example attributions generalize further, capturing more abstract and category-level properties. While higher-ranked images maintain a closer similarity to the test image in terms of critical features like wheels and overall vehicle structure, lower-ranked images deviate more, often representing generalized characteristics of the automobile class. For instance, some lower-ranked images, such as layer 19 rank 8 and layer 20 rank 6, appear visually dissimilar to the test image, yet they retain essential discriminative properties, such as the presence of wheels and a land vehicle's structural form. This suggests that deeper layers prioritize semantic alignment over instance-specific details.

Through these observations of example attributions generated by EBE-DNN, we arrive at a conclusion consistent with \cite{zeiler2014visualizing}: deeper layers seem to generally produce more discriminative features. This affinity demonstrates that the EBE-DNN method through the example attributions offer some insights into ResNet18’s classification mechanism and that it provides a transparent view of the decision-making process.

\subsection{Predictive Performance}\label{sec:results}
\begin{figure} 
\centering
\includegraphics[width=0.6\textwidth]{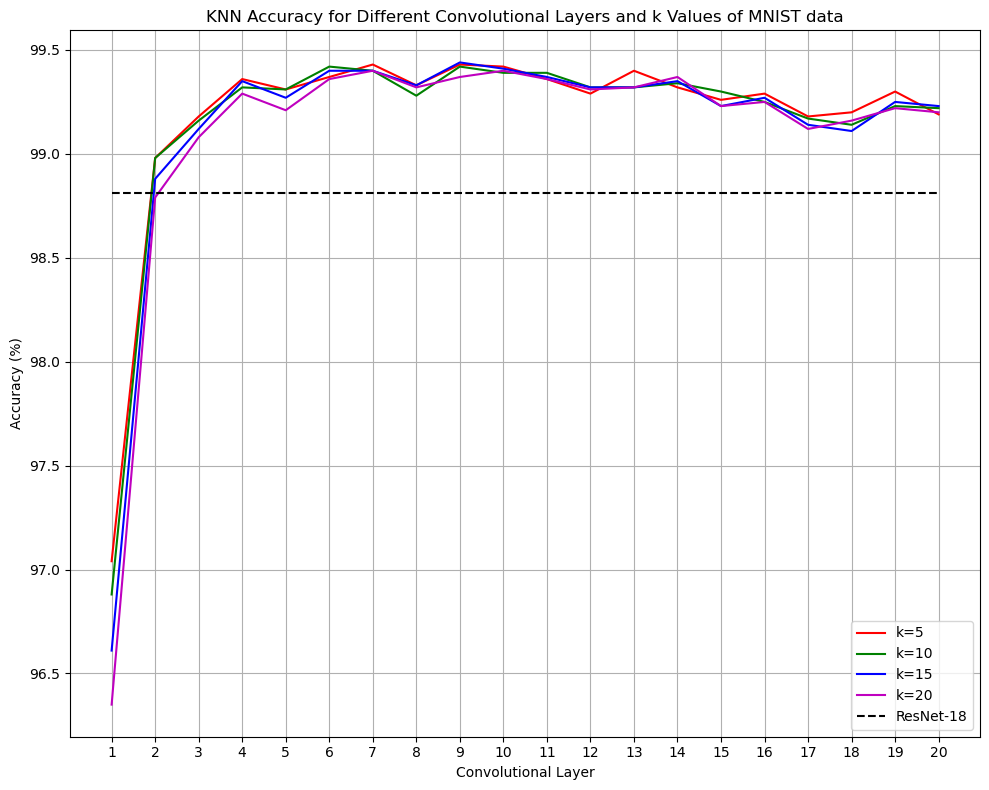}
\caption{Accuracy on MNIST for different layers and number of examples} \label{figmnist}
\end{figure}

In this section, we compare the predictive performance of EBE-DNN to the original DNN, and investigate in particular the effect of the layer to use for the embeddings and the number of examples used in the explanations. 

\subsubsection{Results for MNIST and Fashion-MNIST} 
\begin{figure} 
\centering
\includegraphics[width=0.6\textwidth]{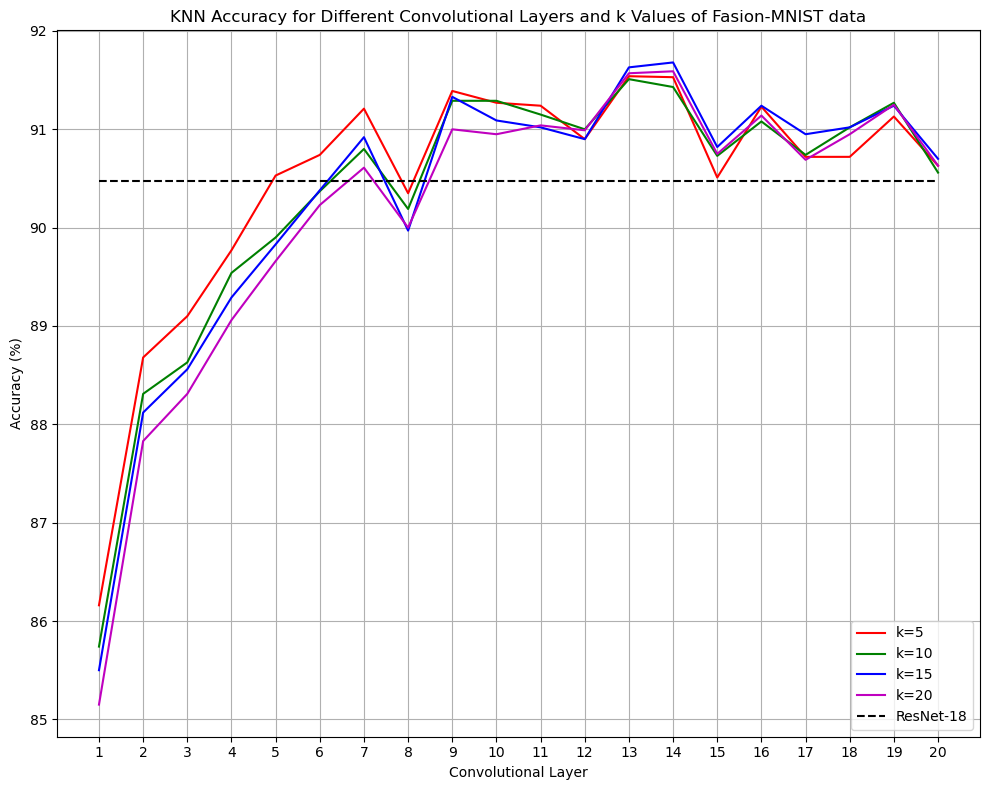}
\caption{Accuracy on Fashion-MNIST for different layers and number of examples} \label{figfashionmnist}
\end{figure}

The predictive performance on the MNIST and Fashion-MNIST datasets are shown in Fig.~\ref{figmnist} and Fig.~\ref{figfashionmnist} for different values of $k$ (number of examples used in the attributions) and the choice of convolutional layers. Notably, despite limiting the number of training examples contributing to a prediction, EBE-DNN does not seem to generally degrade performance compared to the original model. On the contrary, the predictive performance is improved in several cases. For MNIST, improvements are observed when using any layer, except for the top-most, independently of the choice of $k$. For Fashion-MNIST, similar results are observed for layers 9 and below. Somewhat 
surprisingly, the highest accuracies of are observed for when the embeddings are extracted from the middle convolutional layers of ResNet18, rather than the deepest layers, as one could have been expecting. Additionally, the choice of examples ($k$)  has only a minor influence on accuracy compared to the significant impact of the embedding layer selection.

\subsubsection{Results for CIFAR-10} 
\begin{figure}
\centering
\includegraphics[width=0.6\textwidth]{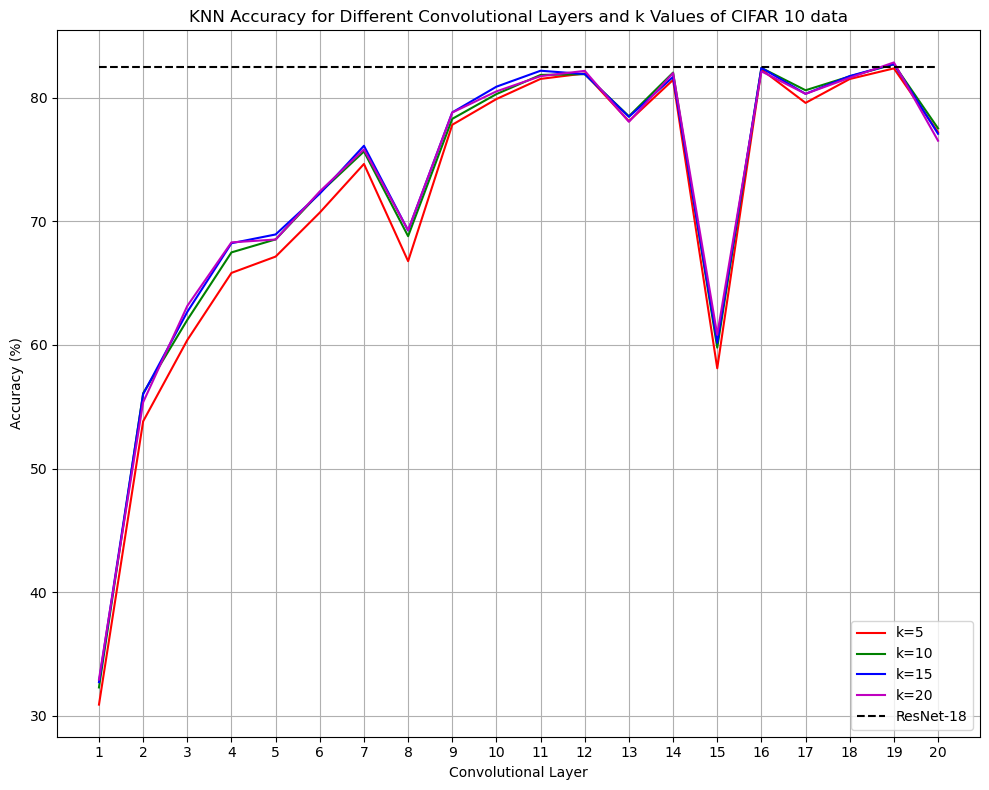}
\caption{Accuracy on CIFAR-10 for different layers and number of examples} \label{figCIFAR}
\end{figure}

The results for the CIFAR-10 dataset are shown in 
Fig.~\ref{figCIFAR}. EBE-DNN achieves accuracy comparable to ResNet18 when the embeddings are extracted from layers 16 and 19 of ResNet18. These layers are particularly effective because their embeddings capture class-specific properties, as demonstrated by the example attributions; most retrieved images belong to the same class as the test image. Interestingly, at layer 15, predictive performance drops significantly, consistent with the attributions presented in Fig.~\ref{figCIFAR-10automobilesim}, where the attributions for this layer included images from multiple other classes. This suggests that embeddings obtained from layer 15 generalize across classes, reducing both attribution quality and accuracy. The results also show that the choice of $k$ has less impact on accuracy compared to layer selection. The results indicate that choosing $k=10$ for this dataset seems to give the best balance of predictive performance and interpretability (as measured by the number of training examples included in the explanations).

\newpage
\section{Concluding Remarks}\label{conclusion}

A novel approach to obtaining example attributions from deep neural networks, called EBE-DNN, has been introduced, which combines KNN with the feature extraction capabilities of DNN to generate example attributions that are both interpretable and aligned with model predictions. By associating test examples with the training data that influence their predictions, EBE-DNN offers a transparent and interpretable framework for emulating the predictions of deep learning models, while maintaining competitive predictive performance.

Our investigation has highlighted the progression of properties captured across ResNet18 layers. Early layers focus on low-level properties like color and texture, which provide surface-level similarities but lack structural information. Middle layers extract distinctive structural features critical for classification, achieving a balance between preserving low-level details and capturing high-level abstractions. Deeper layers generalize to category-level properties, enabling the model to recognize diverse instances within a class but often at the expense of fine-grained details. These insights are consistent with prior work indicating that higher layers produce more discriminative features, demonstrating the interpretability and effectiveness of EBE-DNN in forming example attributions.

In terms of predictive performance, EBE-DNN achieves accuracy comparable to or exceeding ResNet18 across datasets, with middle layers delivering the best results. By leveraging embeddings that capture structural features critical for classification, EBE-DNN balances interpretability and predictive performance. The method performs consistently across different settings of $k$ (number of examples to include in the attributions), with $k=10$ providing a reasonable trade-off between capturing sufficient training examples for accurate predictions and maintaining interpretability by avoiding excessively large sets of retrieved examples.

While effective, EBE-DNN presents potential opportunities for further improvement. One area for future work is addressing the dependence on the choice of convolutional layers, which currently acts as a hyperparameter requiring tuning. Developing automated or adaptive strategies for selecting the most effective layers for attributions and predictions could enhance usability and efficiency. Additionally, while KNN provides interpretable example attributions, it can become computationally expensive for high-dimensional embeddings or large-scale datasets. Future research could explore scalable implementations or alternative similarity-based approaches to reduce computational cost. Other directions concern investigating alternative distance metrics and forming predictions by weighted voting, e.g., using the distances. Another direction for future research is to complement the example attributions with feature attributions, to more clearly determine the factors influencing the model’s predictions. Another direction is a more formal identification and measurement of the concepts used by the model at each layer to make two pictures closer or farther in the latent space. Finally, further exploration is needed to determine whether the EBE-DNN framework is equally effective in other domains, such as text or time-series data, potentially broadening its applicability across machine learning tasks.

\begin{credits}
\subsubsection{\ackname} This project was funded by the Wallenberg AI, Autonomous Systems, and Software Program (WASP).
\end{credits}
%
%
%
\bibliographystyle{splncs04}
\bibliography{references}
\end{document}